\documentclass[11pt]{article}

\usepackage[final]{acl}

\usepackage{times}
\usepackage{latexsym}
\usepackage{amsmath}
\usepackage{amsfonts}

\usepackage{booktabs}
\usepackage{tabularx}
\usepackage{array}

\usepackage{multirow}

\usepackage{algorithm}
\usepackage{algpseudocode}

\usepackage{listings}

\lstdefinestyle{promptstyle}{
    basicstyle=\ttfamily\scriptsize,
    breaklines=true,
    breakatwhitespace=false,
    columns=fullflexible,
    keepspaces=true,
    frame=single,
    xleftmargin=0.5em,
    xrightmargin=0.5em
}

\usepackage{fvextra}
\DefineVerbatimEnvironment{Prompt}{Verbatim}{
  fontsize=\scriptsize,
  breaklines=true,
  breakanywhere=false,
  breaksymbolleft={},
  breaksymbolright={},
  obeytabs=true,
  tabsize=2
}
\usepackage{comment}

\usepackage[T1]{fontenc}

\usepackage[utf8]{inputenc}

\usepackage{microtype}

\usepackage{inconsolata}

\usepackage{graphicx}

\usepackage{paralist}
\usepackage{xspace}
\newcommand{\metric}{$KGain$\xspace}
\newcommand{\metricbold}{\textbf{\textit{KGain}}\xspace}
\newcommand{\metricfull}{{\sc KnowledgeGain}\xspace}

\newcommand{\best}[2]{$\mathbf{#1}{\pm}#2$}

\title{{\sc \textbf{KnowledgeGain}}: Evaluating and Optimizing Science News Generation for Reader Learning}

\author{Dominik So\'os \\
  Old Dominion University \\
  Norfolk, Virginia, USA \\\And
  Meng Jiang \\
  University of Notre Dame \\
  Notre Dame, Indiana, USA \\\And
  Jian Wu \\
  Old Dominion University \\
  Norfolk, Virginia, USA \\ 
}

\begin{document}
\maketitle
\begin{abstract}
Science news is an important medium to communicate discoveries between the research communities and the public. 
Yet, most metrics for generated or summarized text evaluate semantic similarity and factual consistency, do not measure how much knowledge readers learn from the news.
We introduce \metricfull, a metric that evaluates the quality of science news by measuring how much knowledge readers gained after reading it. 
To evaluate the metric, we first performed a controlled human study and showed that the metric successfully captures the differential knowledge gained by human readers reading different types of science media. 
The data allowed us to calibrate a prompt-only LLM reader simulator. 
We use it to rank and filter candidate articles before human evaluation. 
A second human study shows that articles selected with this simulator improve post-reading accuracy and normalized \metricfull over a strong generation baseline.
Our work is a step toward generating science news that better meet the knowledge and comprehension goals of Bloom's Taxonomy. 
\end{abstract}


\section{Introduction}
Science is often regarded as a common inheritance of humanity \citep{pasteur1939toast}. 
However, scientific publications are often written in highly specialized language that the vast majority of the general public cannot readily understand. 
Even with the growth of open-access in removing paywalls, the gap between scientific knowledge and public comprehension \citep{suber2012open,bucchi2008deficits} still exists and is widening. 
Inspired by this need, computational linguists and NLP researchers aim to develop AI-driven methods for automatically generating science news from scientific papers, making science not only open in access, but open in language.
However, how to effectively evaluate AI-generated science news computationally remains a foundational question.
Standard evaluation metrics for text generation and summarization such as ROUGE \citep{lin2004rouge}, BERTScore \citep{zhang2020bertscore}, and BLEURT \citep{sellam2020bleurt} primarily measure the lexical or semantic similarity against references. 
More recent metrics focus on factual consistency between a summary and its source using question answering or natural language inference \citep{kryscinski2020factcc,fabbri2022qafacteval,scialom2021questeval,deutsch2021qaeval}. 
These metrics either assume a ground truth exists or the evaluation is based on a reference text.
Reader learning, which serves as an important goal of science news, is not incorporated. 

This creates a significant gap especially important in science news generation. 
Early work on lay summarization of scientific text improves the comprehension of non-experts \citep{stoll2022plos,maurer2021lessons,anderson2022comparison}.
However, science news readers cover a broader spectrum of audiences, requiring more public oriented language and faithfully preserving scientific facts, so readers could learn authentic knowledge without strong domain background \citep{shulman2024simplicity,yeung2018neuro}.
Therefore, a fundamental question to bridge this gap is: \emph{How to evaluate the learning usefulness of science news and compare it with other types of science communications by measuring the knowledge gained by readers?}

\begin{figure*}
    \centering
    \includegraphics[width=0.84\textwidth]{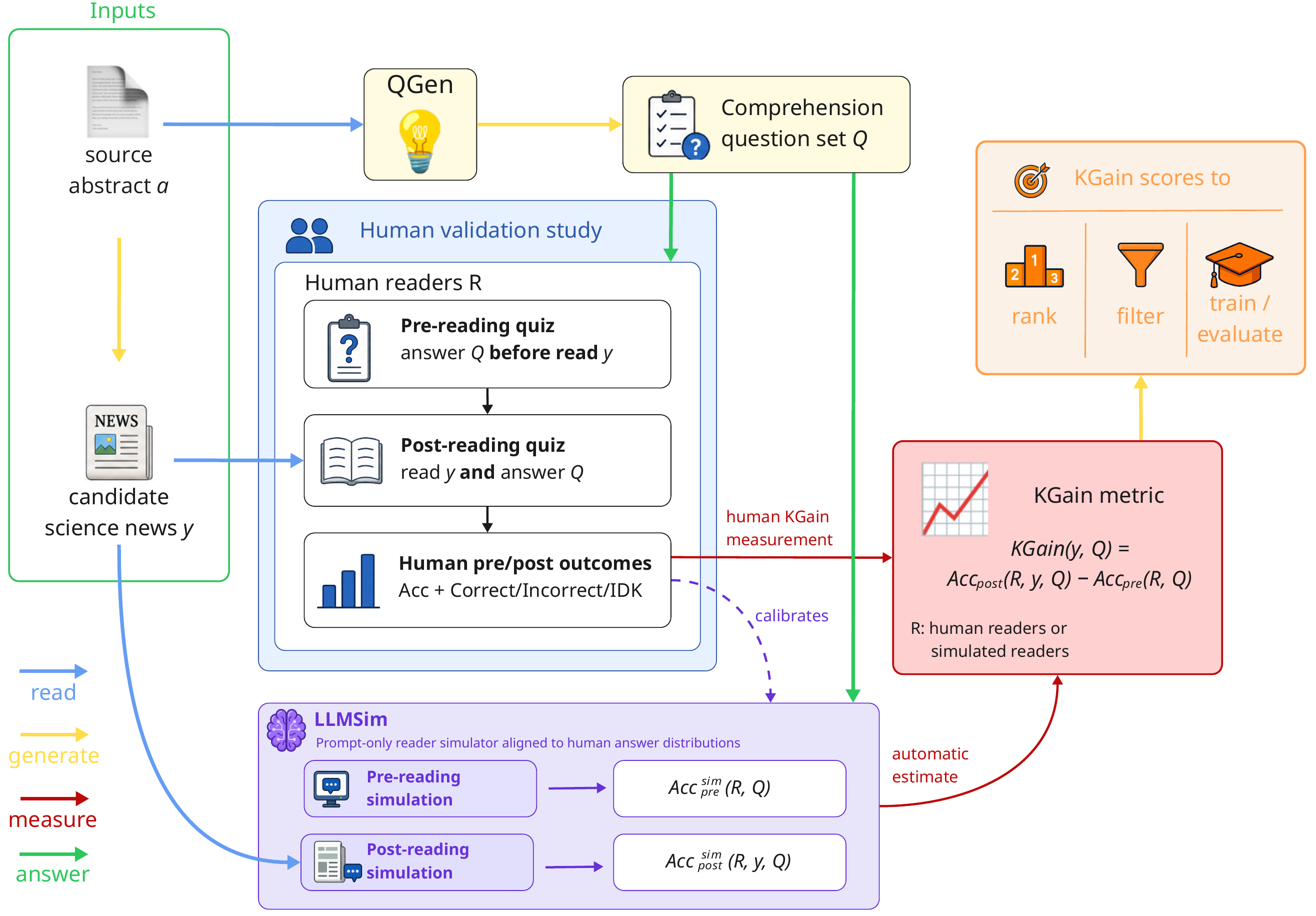}
    \caption{Overview of the pipeline that estimates \metric with human-calibrated {\sc LLMSim}. Given a source abstract $a$, {\sc QGen} generates a comprehension question set $Q$. Human readers answer the same questions before and after reading a candidate science news article $y$, producing distributions over Correct, Incorrect, and "I do not know" (IDK). These distributions calibrate {\sc LLMSim}, which estimates pre/post accuracy. The resulting \metric is used to score and rank articles, filter training data, and train or evaluate science news generation.}
    \label{fig:flow}
\end{figure*}

Knowledge assessment has been extensively studied in cognition and educational theory. 
Foundational work has been established by Bloom's taxonomy \citep{bloom1956taxonomy}, revised by \newcite{anderson2001taxonomy} about the levels of cognitive learning, and the concept of learning gain \citep{hake1998normalized}, formally defined as $g = \frac{post - pre}{max - pre}$, in which $pre$ and $post$ are the scores before and after an intervention. 
Inspired by these two theories, we propose a novel computational metric \metricfull, denoted by \metric, to measure the usefulness of science news based on how much knowledge learned by readers.
It leverages questions as proxies to measure the knowledge corresponding to the knowledge and comprehension layers in Bloom's Taxonomy. 
Though motivated by science news, \metric intuitively is applicable to other types of science communications such as abstracts and tweets/threads. 


We first validate \metric in a controlled human study and demonstrate that it captures meaningful differences of knowledge gained through reading three typical communication types. 
To scale reader-centric evaluation beyond expensive human studies, we develop {\sc LLMSim}, a reader simulator model to simulate the reading and question-answering process of human readers. 
We demonstrate that {\sc LLMSim} provides a scalable estimation of \metric that allows us to qualitatively evaluate science news which can be further used for filtering automatically generated news. 
Figure~\ref{fig:flow} summarizes the pipeline.

We perform rigorous evaluation of \metric.
First, we test whether \metric distinguishes knowledge learned from three types of science communications: abstracts, science news, and tweets. 
Second, we evaluate whether \textsc{LLMSim} can approximate human pre/post distributions and approximates aggregate \metric.
Third, we use \metric estimated by {\sc LLMSim} to rank and filter science news, which is used for training  science-news generators, and evaluate the resulting news with human judgments. 
Finally, we compare \metric with standard similarity metrics and LLM-judge scores to test whether \metric captures orthogonal dimensions not captured by standard metrics.

This paper makes three major contributions:
\begin{compactitem}
    \item \textbf{A reader-centered evaluation metric.} We introduce {\sc KnowledgeGain}, a pre/post comprehension metric for evaluating how useful science communication is for reader learning and apply it to science news evaluation and generation. 

    \item \textbf{A calibrated simulated reader.} We develop {\sc LLMSim}, a prompt-only LLM simulator aligned with human answer outcome distributions, providing a scalable tool to estimate \metric for generated science news. 

    \item \textbf{Science news dataset.} We construct a 2,747-instance abstract--QA--news corpus containing \metric-filtered news as a baseline for future science news generation models.
\end{compactitem}

\section{Related Work}
\label{sec:related}

\subsection{Evaluating Generated Text}
Automatic evaluation of generated text has traditionally relied on n-gram overlap such as ROUGE, BLEU, and METEOR, or learned similarity like BERTScore and BLEURT \citep{lin2004rouge,papineni2002bleu,banerjee2005meteor,zhang2020bertscore,sellam2020bleurt}.
Although useful, these metrics often correlate unevenly with human judgments in terms of quality dimensions and datasets. 
To move beyond measuring token overlap, QA- and NLI-based evaluate factual consistency, and content preservation, including FEQA \citep{durmus2020feqa}, QAEval \citep{deutsch2021qaeval}, QuestEval \citep{scialom2021questeval}, FactCC \citep{kryscinski2020factcc}, SummaC \citep{laban2022summac}, and QAFactEval \citep{fabbri2022qafacteval}.
Benchmarks like FRANK \citep{pagnoni2021frank}, SummEval \citep{fabbri2021summeval}, and toolkits such as SacreROUGE \citep{deutsch2020sacrerouge} enabled large-scale meta-evaluation.
Surveys continue to refine best practices for factuality evaluation \citep{luo2024factual}.
These metrics assess textual similarity, factual consistency, or preference, but they do not directly measure how much a reader learned from the text.

Recent work uses LLMs as reference-free evaluators, often called LLM-as-a-judge, with promising agreement to humans in open-ended tasks \citep{liu2023g,zheng2023judging}.
However, LLMs can be unreliable for scoring the aspects of news and pairwise comparisons and do not provide quantitative measurements of the usefulness between two news articles. 

\subsection{Learning-Centered Evaluation}
Pre/post testing is a long-standing paradigm to quantify learning, including the normalized gain popularized by \citet{hake1998normalized}.
In information retrieval, \emph{search-as-learning} models change in user knowledge states over a session, typically using pre/post question-answering accuracy \citep{yu2021topic,ghafourian2021kair,elzein2023sal,nasser2024rulkkg,tibau2023accounting}.
We adopt an interpretable variant for science communication: raw knowledge gain, defined as post-reading accuracy minus pre-reading accuracy. 

\subsection{LLMs as Simulated Readers}
The main obstacle in using LLMs as reader simulators is \textbf{hyper-accuracy distortion}: even when instructed to act like an average student, an LLM may still infer the correct answer from linguistic cues and become far more accurate and less uncertain than human readers \citep{aher2023using,amirova2024framework}. 
This is closely related to recent findings that LLMs may better express human choice distributions when asked to verbalize probabilities than when sampled for a single forced-choice answer \citep{meister2025benchmarking,xie2025distributional}.

A growing line of work therefore frames simulation as \emph{distribution matching} rather than pointwise prediction.
\emph{Mixture-of-persona} methods represent a population with multiple personas rather than one ``average human'' prompt \citep{xie2025distributional,bui2025mixture}, motivating our use of persona clusters sampled according to their empirical frequency in the human validation study.

Personas alone are not enough if the model can rederive answers from the full article. 
A more realistic reader simulator should answer from memory, not by  searching the original text, which motivates our \emph{memory bottleneck} and connects to cognitive accounts of gist and verbatim memory \citep{reyna2016fuzzy,reyna2012new}.
We model \emph{abstention} directly because \metric depends on whether readers move from IDK or from an incorrect belief to the correct answer \citep{wen2025know,madhusudhan2025llms,kapoor2024large}.
To avoid overconfident answers, we use \emph{verbalized sampling} over answer options including IDK \citep{aher2023using,kapoor2024large,zhang2025verbalized}.

\section{{\sc \textbf{KnowledgeGain}}}
\label{sec:kgain}
We define {\sc KnowledgeGain} as the change in a reader's QA accuracy after exposure to a science communication artifact.
Given an artifact $y$, a reader $r$, and a set of comprehension questions $Q$, the pre-reading QA accuracy is:
\begin{equation}
\mathrm{Acc}_{pre}(r,Q) =
\frac{1}{|Q|}\sum_{q \in Q}
\mathbb{I}[\hat{a}^{pre}_{r,q} = a_q^*],
\end{equation}
and the post-reading accuracy is:
\begin{equation}
\mathrm{Acc}_{post}(r,y,Q) =
\frac{1}{|Q|}\sum_{q \in Q}
\mathbb{I}[\hat{a}^{post}_{r,q} = a_q^*],
\end{equation}
where $a_q^*$ is the correct answer to question $q$. 
The accuracy-based \metricfull is then:
\begin{equation}
{KGain}(r,y,Q) =
\mathrm{Acc}_{post}(r,y,Q) -
\mathrm{Acc}_{pre}(r,Q).
\end{equation}

Since the raw \metric can be affected by ceiling effects when pre-reading accuracy is high, we compute normalized \metric as a robustness check:
\begin{equation}
g =
\frac{\mathrm{Acc}_{post} - \mathrm{Acc}_{pre}}
{1 - \mathrm{Acc}_{pre}}.
\end{equation}

The mean \metricfull of a population of readers $R$ is then:
\begin{equation}
{KGain}(y,Q) =
\frac{1}{|R|}\sum_{r \in R}
{KGain}(r,y,Q).
\end{equation}

\subsection{Question Set Construction}
\label{sec:question_set_construction}

Given a reading document, $K$ multiple-choice comprehension questions are designed to first measure the \emph{Knowledge} and \emph{Comprehension} layers in Bloom's taxonomy \citep{bloom1956taxonomy}. 
To facilitate the QA process, we developed {\sc QGen} that produces questions based on the information source.
To meet the goals of measuring the knowledge levels in Bloom's Taxonomy, we require all questions to be directly answered (knowledge) or the answers are inferrable from the information source (comprehension).  
We define a question as answerable if the correct option is either directly stated in the source or can be inferred from one or more source statements without requiring outside knowledge. 
Questions whose answers cannot be recovered from the source, have multiple plausible correct options, or depend on assumptions are replaced.

To incrementally measure the capability of knowledge gain, we design 6 questions in three tiers of difficulty, each mapping to a layer of Bloom's taxonomy.
Q1--2 are extractive true/false questions targeting answers explicitly present in the source;
Q3--4 are extractive multiple choice questions with distractors; and
Q5--6 are inferential multiple choice questions requiring the readers to synthesize information not explicitly stated but can be inferred from the source. 
Together, these question tiers cover factual recall and basic comprehension in Bloom's taxonomy. 
Each question includes an explicit ``I do not know'' (IDK) option, scored as incorrect for accuracy but preserved as an abstention outcome. 
This design allows us to distinguish uncertainty from incorrect beliefs. 

\subsection{Human-based Measurement}
Human-based \metric can be measured with a pre/post protocol by human readers. 
Readers first answer the question set using background knowledge, then read an artifact, and finally answer the same questions without returning to the artifact. 

\subsection{Automatic Measurement}
It is too expensive to apply humans for every generated article during data construction and model development.
We therefore introduce {\sc LLMSim}, a prompt-only LLM procedure that approximates human behavior and estimates \metric.
{\sc LLMSim} is not intended to replace human evaluation.
Instead, it provides a scalable proxy for scoring large numbers of generated articles before final human validation.
{\sc LLMSim} simulates readers in the same stages as the human study: answer from background knowledge, read the artifact, form a short memory trace, and answer again without returning to the text. 

\paragraph{Mixture-of-personas.}
We model the reader population as a mixture-of-persona clusters derived from the human validation study.
We first inspect participants' response patterns, including baseline knowledge, tendency to abstain, and post-reading improvement, and defined five reader personas that capture recurring behaviors. 
Each human participant is assigned to one persona cluster, and simulated readers are sampled from these clusters according to their empirical frequency in the validation study. 
The exact prompts and their frequencies can be found in Appendix~\ref{app:personas}.

\paragraph{Memory bottleneck.}
To prevent direct extraction from the full text, the simulator does not answer post-reading questions while viewing the original article. 
Instead, it first produces a short memory trace summarizing what the reader retained, then answers the questions using only that trace. 
We use a dual-trace variant that produces two possible recollections and a probability for each trace, which captures variability in what readers may remember.

\paragraph{Abstention.} 
Because each question includes an IDK option, {\sc LLMSim} can abstain when it lacks background knowledge, evidence or memory trace to an answer. 
We score IDK as incorrect for accuracy, but preserve it as a separate outcome during calibration because moving from IDK to Correct means the article filled a knowledge gap, and moving from Incorrect to Correct means the article corrected a misconception.

\paragraph{Verbalized sampling.}
To avoid mode collapse and inhumanly confident behavior, the simulator does not directly output a single option. 
Instead, it verbalizes a categorical distribution over answer options including IDK.
We sample one answer from this distribution. 
This produces more human-like variation across simulated reader while avoiding instructions to intentionally answer incorrectly.

\paragraph{Calibration.}
We calibrate LLMSim by matching human outcome distributions from the validation study. 
For each topic $i$, question $j$, and condition $m \in \{\text{Pre},\text{News}, \text{Abstract}, \text{Tweet}\}$, let $P_{i,j}$ denote the empirical human distribution over the three outcomes. 
We tune simulator parameter, $\theta$, to minimize KL divergence, a standard distributional mismatch, between the human and simulated distributions:
\begin{equation}
\theta_m^* =
\arg\min_{\theta}
\sum_{i,j}
D_{\mathrm{KL}}
\left(
P_{i,j,m} \,\|\, Q_{\theta,i,j,m}
\right).
\end{equation}

Matching accuracy alone would collapse IDK and incorrect answers into the same category. 
Instead, this distributional objective preserves IDK as a separate calibration target, allowing it to model both uncertainty and wrong beliefs. 
The full algorithm is shown in Appendix~\ref{app:algorithm}.

\paragraph{Simulated \metricbold Scoring.}

After calibration, we use {\sc LLMSim} to estimate \metric for generated articles.
For each article, {\sc LLMSim} samples simulated readers, computes pre-reading and post-reading accuracy over the associated question set.

\metric measures targeted comprehension gain for a fixed question set $Q$, not scientific understanding in general. 
Because $Q$ is generated before evaluation and screened for answerability and uniqueness, \metric is most appropriate for evaluating whether an article communicates the source claims needed to answer factual and inferential comprehension questions.

\section{Evaluation}
\label{sec:evaluation}
We evaluate \metricfull in four stages: first as a metric across different communication types using human readers, then we demonstrate that \metric can be automatically estimated with {\sc LLMSim}, then as an optimization signal for science news generation and finally we verify that \metric optimized news let human readers learn more knowledge. 

\begin{figure}[ht!]
    \centering
    \includegraphics[width=\columnwidth]{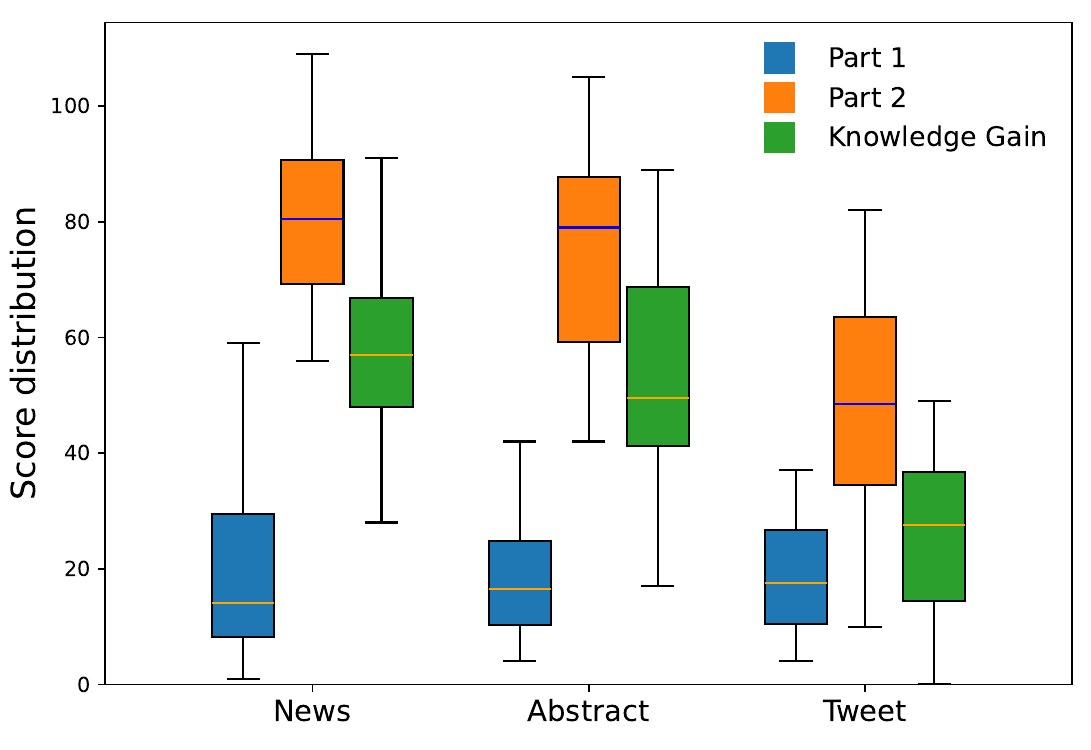}
    \caption{Score distributions for Part 1 (pre-reading), Part 2 (post-reading), and the resulting \metricfull across the three media types}
    \label{fig:kgain}
\end{figure}

\subsection{Human-based \metricbold}
In this experiment, we calculate \metric by letting human readers answer questions before and after reading three types of science communications. 
We aim at testing two hypotheses: science news yields \metric comparable to abstracts, and science news yields higher \metric than tweets/threads.

\paragraph{Experimental Setup.}
We validate \metric on 30 research topics, each represented by a paper abstract, a professionally written science news\footnote{All human-written news articles are from sciencealert.com}, and a tweet/thread from X.com. 
We automatically collected the news that linked to papers, manually verified, and used those paper links to find promotional posts or threads. 
The resulting topic set covers six broad science areas in physics, nature, space, health, humans, and technology.
We then use {\sc QGen} to generate six comprehension questions from the abstract, yielding 180 total questions manually verified following the criteria in $\S$~\ref{sec:question_set_construction}.

We recruited 30 STEM undergraduates and randomly divided them into three 10 person groups with a balanced GPA distribution.
The 30 topics were split into three blocks of 10 topics and each group evaluated one artifact type per block, as shown in Table~\ref{tab:crossover_design}.

\begin{table}[t]
\centering
\footnotesize
\setlength{\tabcolsep}{6pt}
\begin{tabular}{lccc}
\toprule
Group & Block 1 & Block 2 & Block 3 \\
 & D1--D10 & D11--D20 & D21--D30 \\
\midrule
G1 & News & Abstract & Tweet/Thread \\
G2 & Abstract & Tweet/Thread & News \\
G3 & Tweet/Thread & News & Abstract \\
\bottomrule
\end{tabular}
\caption{Crossover block design for the human validation study. Each group evaluates all three artifact types, and each topic block is evaluated under all three artifact types across groups.}
\label{tab:crossover_design}
\end{table}

\paragraph{Results.}
The results of our human study are used for testing the two hypotheses above.

\textbf{News vs. Abstracts (H1).}
As shown in Figure~\ref{fig:kgain}, background knowledge was consistently low across all groups (median scores $<20$). 
Post-reading scores show that News articles achieved a median score of approximately $80$, slightly outperforming Abstracts ($\approx78$). Most importantly, the \metric distribution for News (median $\approx60$) is comparable to, and shows less variance than, that of Abstracts (median $\approx50$). 
This supports the hypothesis that science news achieves comparable to or higher \metric than paper abstracts, which is consistent with the intuition. 
Most undergraduate students have not professionally trained on reading scientific papers. 
However, by reading science news, these students gain comparable or even more knowledge than reading paper abstracts, which is reflected in \metric. 

\textbf{News vs Tweets (H2).}
In contrast, Tweets yielded significantly lower \metric compared to both News and Abstracts. 
While Tweets required the least reading time (Figure~\ref{fig:reading_times} in Appendix~\ref{app:reading_time}), they produced substantially lower knowledge gain, supporting H2.

\paragraph{Implications.}
The results of this experiment demonstrate that although news articles required the longest reading time, they also produced the largest absolute gains in knowledge. 
This suggests that the additional reading time is associated with greater knowledge gain, compared with abstracts and tweets/threads and \metric provides a useful way to distinguish the usefulness of learning of different types of science communications. 
A mixed-effects robustness check confirms the pattern ($p<0.001$): News and Abstract yield larger gains than Tweet, while News and Abstract are not significantly different; full results are in Appendix~\ref{app:human_validation_mixed_model}.

\subsection{Automatic KGain Estimation}
\label{sec:automatic_kgain_estimation}
We evaluate whether {\sc LLMSim}, defined in $\S$~\ref{sec:kgain}, provides a reasonable automatic proxy for human pre/post learning. 

\paragraph{Alignment with human readers.}
From Table~\ref{tab:llmsim_condition_alignment}, we can observe that {\sc LLMSim} aligns reasonably well in News and Abstract aggregate distributions, but Tweet remains harder. 
This makes sense because tweets are short, under-specified and more prone for ambiguity. 
The aggregate distributions show that {\sc LLMSim} aligns closely with human post-reading behavior for News and Abstract conditions, especially abstention rates.
The main mismatches are on ``Pre,'' where the model remains more knowledgeable than humans, and ``Tweet,'' where the simulator over-abstains.

\begin{table}[t]
\centering
\small
\setlength{\tabcolsep}{5pt}
\begin{tabular}{lrrr}
\toprule
Condition & KL $\downarrow$ & Correct MAE $\downarrow$ & IDK MAE $\downarrow$ \\
\midrule
Pre & 0.090 & 0.151 & 0.193 \\
News & 0.021 & 0.099 & 0.009 \\
Abstract & 0.007 & 0.059 & 0.010 \\
Tweet & 0.119 & 0.068 & 0.224 \\
\bottomrule
\end{tabular}
\caption{Alignment between human and full {\sc LLMSim} answer-outcome distributions by condition. KL measures aggregate alignment over Correct/Incorrect/IDK outcomes. Correct mean absolute error (MAE) and IDK MAE measure calibration of correctness and abstention rates. Lower values are better.}
\label{tab:llmsim_condition_alignment}
\end{table}

\paragraph{Ablation.}
LLMs are not naturally good human readers: simulator design must be careful.
The results in Table~\ref{tab:llmsim_ablation} show that the main challenge is not simply prompting an LLM to act like a student. 
Direct prompting produces a hyper-accuracy simulator with almost no uncertainty. 
The most important design choices are therefore mechanisms that mimic human-like uncertainty.

\begin{table*}[ht]
\centering
\small
\setlength{\tabcolsep}{5pt}
\begin{tabular}{lrrrrr}
\toprule
Simulator variant  & Global KL $\downarrow$ 
& Item KL $\downarrow$  & \metric corr. $\uparrow$ & Correct MAE $\downarrow$  & IDK MAE $\downarrow$ \\
\midrule
Direct answer & 4.953 & 5.099 & -0.063 & 0.611 & 0.254 \\
+ Persona mixture & 4.897  & 4.975  & -0.044  & 0.601  & 0.254 \\
+ Memory bottleneck  & 4.613  & 4.183  & 0.065  & 0.526  & 0.254 \\
+ IDK option  & 0.381 & 3.653 & \textbf{0.272} & 0.395 & 0.306 \\
+ Verbalized sampling  & \textbf{0.023}  & \textbf{1.934} & 0.263  & \textbf{0.329}  & \textbf{0.270} \\
\bottomrule
\end{tabular}
\caption{Ablation of {\sc LLMSim} components against human pre/post answer-outcome distributions. Direct answering substantially overestimates human correctness and never abstains. Adding an explicit IDK option and verbalized sampling sharply improves distributional alignment. Bold numbers indicate the best numerical values.}
\label{tab:llmsim_ablation}
\end{table*}

Table~\ref{tab:llmsim_ablation} shows that naive LLM answering is a poor proxy for human learning behavior: it is hyper-accurate and never abstains. 
Persona prompting alone has little effect, and the memory bottleneck reduces over-accuracy but does not capture uncertainty. 
The largest gains come from modeling uncertainty directly through the IDK option and verbalized sampling, which sharply reduce global KL and bring aggregate correctness close to the human rate. 
Item-level KL remains higher than global KL, so we treat {\sc LLMSim} as a scalable proxy for relative \metric comparison rather than a replacement for human evaluation.


\subsection{KGain optimized Science News Generation}
\label{sec:kgain_generation}
In the above experiments, we showed that \metric can be used to evaluate and compare the usefulness of different kinds of science communications, including science news, and that it was possible to design an automatic evaluator for a given science news. 
However, we have not verified that filtering generated news with simulated \metric actually improves article quality or whether it helps readers learn more than they would from a strong baseline news generator. 

\paragraph{Setup.}
To fine-tune a strong science news generator, we construct a corpus of 2,747 instances. 
Each contains a scientific abstract, comprehension questions generated using {\sc QGen} with ground truth answers, multiple candidate news articles, and  the best news articles ranked by \metric. 
The candidate articles are generated by a teacher model using direct prompting. 
An open-source model scores each candidate for factual accuracy, completeness, relevance, clarity, and an estimated \metric given the QA set for each abstract. 
This corpus is used to construct both unfiltered and \metric-filtered SFT training sets for fine-tuning the generator. 

We evaluate generated articles on 300 held-out abstracts using {\sc LLMSim}. 
Given abstract $a$, we prompt a generator model to compose science news with a headline-first, multi-paragraph structure, journalistic tone.
The generator is \emph{not} allowed to access any reference news article.
We sample $C$ candidates per abstract to form a candidate set $\{y_1,\dots,y_C\}$.

We use {\sc LLMSim} to score candidate articles by simulated \metric and construct the \metric-filtered SFT dataset by selecting candidates whose simulated \metric scores are positive. 
We fine-tune a news generator using the data selected.
Importantly, {\sc LLMSim} is used only as a development time filtering signal, our final claims are based on blinded human evaluations and human learning measures.
We then compare our model against 3 baseline models: 
Baseline 1, a zero-shot news generator; Baseline 2, an agentic news generator; and Baseline 3, human-written science news collected from ScienceAlert for the corresponding paper abstract. 
We found Baseline 2 to be the strongest based on Figure~\ref{fig:llm_judge} in Appendix~\ref{app:pointwise-judge}.
All news generators are based on \texttt{Qwen3-4B-Instruct-2507}\footnote{We use the same LLM for all automatic generators. We do not test closed-source models, as a stronger backbone because they would likely improve all automatic systems without changing the relative comparison.}, and generator names are anonymized during human and LLM evaluations.

\begin{table}[ht]
\centering
\tiny
\setlength{\tabcolsep}{2.2pt}
\begin{tabular}{lrrrrr}
\toprule
Generator & Accuracy & Completeness & Relevance & Clarity & Mean \\
\midrule
Agentic & $3.34{\pm}0.67$ & $4.30{\pm}0.57$ & $4.34{\pm}0.82$ & $4.06{\pm}0.68$ & $4.01{\pm}0.63$ \\
Ours & \best{4.24}{0.34} & \best{4.57}{0.18} & \best{4.50}{0.43} & \best{4.08}{0.56} & \best{4.35}{0.28} \\
\bottomrule
\end{tabular}
\caption{Pointwise evaluation of generated articles. Scores are on a 1--5 scale, with higher values indicating better quality. Values are means with 95\% confidence intervals computed over respondent-level means. Bold values indicate the best score in each column.}
\label{tab:human_pointwise_generated}
\end{table}

\paragraph{Pointwise evaluation.}
We first check whether \metric optimization preserves basic article quality.
Participants rated blinded articles on a 5-point Likert scale for accuracy, completeness, relevance, and clarity while having access to the abstract. 
We compare our \metric-optimized generator against the strongest baseline. 
Table~\ref{tab:human_pointwise_generated} shows that our system receives a higher mean score with the largest difference on accuracy. 
This suggests that optimizing for learning gain does not come at the cost of basic quality.

\paragraph{Pairwise evaluation.}
We also test whether readers prefer the \metric-optimized articles directly.
Participants chose between two blinded articles given the same abstract. 
The \metric-optimized system was preferred over the agentic baseline in 87.0\% of comparisons (Wilson 95\% CI: [79.0, 92.2]; exact binomial test against chance, $p<.001$).
This provides a direct preference check that is independent of the simulated filtering score.

\paragraph{Human \metricbold evaluation.}
Finally, we directly evaluate whether \metric-optimized generation improves reader learning.
Participants answered the same comprehension questions before and after reading generated articles, allowing us to compute pre-reading accuracy, post-reading accuracy, raw \metric, and normalized \metric.
In Table~\ref{tab:human_kgain_generated}, the optimized model significantly improves normalized \metric over the baseline ($p=.048$) and also improves post-reading accuracy.
Results are averaged over 20 topics and 640 participant--article observations.
A mixed-effects model confirms the normalized-gain effect ($p=.015$).
Full results are in Appendix~\ref{app:generated_human_kgain_models}.
Raw \metric shows a nonsignificant positive trend.
Because normalized \metric accounts for the remaining room for improvement after pre-reading accuracy, we use it as the primary human learning measure in the correlation analysis below.
At the system level, {\sc LLMSim} predicts the same aggregate direction as the human study.
However, simulated and human per-topic normalized \metric gains are weakly correlated, so we use {\sc LLMSim} as a development-time proxy rather than a replacement for human pre/post evaluation.

\begin{table}[t]
\centering
\small
\setlength{\tabcolsep}{2.5pt}
\begin{tabular}{lrrrrr}
\toprule
Generator & Pre & Post & {\sc KGain} & Human $g$ & Sim. $g$ \\
\midrule
Agentic & 0.410 & 0.855 & 0.446 & 0.719 & 0.330 \\
Ours & 0.424 & \textbf{0.881} & \textbf{0.457} & \textbf{0.777} & \textbf{0.378} \\
\midrule
$\Delta$ & +0.015 & +0.026 & +0.011 & +0.058 & +0.048 \\
Wilcoxon $p$ & 0.614 & 0.032 & 0.695 & 0.048 & 0.040 \\
\bottomrule
\end{tabular}
\caption{Human pre/post \metric evaluation on generated articles. Human $g$ denotes normalized \metric; Sim. $g$ denotes {\sc LLMSim} normalized \metric. Bold values indicate the better generator result for post-reading and gain metrics.}
\label{tab:human_kgain_generated}
\end{table}

\subsection{\metricbold and Existing Evaluation Metrics}
\label{sec:existing_metrics}

We compute standard reference-based generation metrics between generated news and the paper abstract:
ROUGE-1/2/L \citep{lin2004rouge}, BLEU \citep{papineni2002bleu}, and BERTScore \citep{zhang2020bertscore}.
These metrics measure surface/semantic similarity to a reference, but do not directly measure reader learning.
We therefore use them to quantify the relationship between \metric and traditional similarity-based evaluation.

\paragraph{Correlation analysis.}
To test whether existing metrics are predictive of reader learning, we compute Spearman and Kendall rank correlations between \metric and common reference-based metrics. 
We report both (i) news generated from the same abstracts and (ii) news articles correlations pooled across abstracts, using bootstrap confidence intervals. 
A weak correlation supports the claim that \metric captures an orthogonal dimension.

To assess the significance of \metric differences, we use paired nonparametric tests (Wilcoxon signed-rank) across topics. 
Finally, we compute Spearman and Kendall rank correlations and between \metric and reference-based metrics to determine if traditional evaluation captures the ``instructional'' dimension of science communication.


Table~\ref{tab:metrics_vs_human_kgain} shows that standard automatic metrics correlate only weakly with normalized \metric. 
Source-overlap metrics such as ROUGE-1, ROUGE-2, and BLEU show near-zero association with human normalized \metric. 
BERTScore is weakly associated with \metric, suggesting that semantic similarity to the source abstract is not sufficient to predict what readers learn. 
LLM judge dimensions such as accuracy, completeness, relevance, and clarity likewise show weak correlations.


\begin{table}[t]
\centering
\small
\setlength{\tabcolsep}{4pt}
\begin{tabular}{lrr}
\toprule
Metric & Spearman $\rho$ & Kendall $\tau_b$ \\
\midrule
ROUGE-1 & 0.177 & 0.121 \\
ROUGE-2 & 0.296 & 0.208 \\
ROUGE-L & 0.205 & 0.138 \\
BLEU & 0.267 & 0.167 \\
BERTScore & 0.232 & 0.149 \\
Judge accuracy & 0.291 & 0.240 \\
Judge completeness & -0.016 & -0.012 \\
Judge relevance & 0.033 & 0.028 \\
Judge clarity & 0.089 & 0.072 \\
Judge exp. \metric & 0.168 & 0.139 \\
\bottomrule
\end{tabular}
\caption{Rank correlations between automatic metrics and human normalized \metric. Spearman's $\rho$ and Kendall's $\tau_b$ measure whether automatic metrics preserve the ordering of articles by reader learning.}
\label{tab:metrics_vs_human_kgain}
\end{table}


\paragraph{LLM-judge comparison.}
We apply the same pointwise rubric and pairwise comparison protocol with an independent LLM judge to test whether judge-based evaluation reproduces human judgments. 
LLM pointwise scores correlate weakly with human mean scores overall ($r=.19$).
Full plots and per-dimension correlations are reported in Appendix~\ref{app:human_llm_alignment}. 
This motivates direct reader-centered evaluation with \metric rather than relying only on generic judge ratings.


\paragraph{Implications.}
These results indicate that \metric captures a distinct reader-centered signal that is not reducible to lexical overlap, semantic similarity, or generic quality ratings.

\section{Conclusion and Future Work}
We introduced {\sc KnowledgeGain}, a reader-centered metric for evaluating science news by measuring what readers learn. 
We validated it with human readers, calibrated {\sc LLMSim} as a scalable proxy, and used simulated \metric to filter training data for generation. 
Human evaluations show that the resulting articles improve normalized learning gains over a strong agentic baseline while preserving article quality. 
We leave direct reward-model optimization to future work.

\section{Limitations}
Our human studies use controlled participant pools and do not fully represent the broader public audience for science news. Future work should evaluate \metricfull with more diverse readers, including non-STEM audiences.

\metricfull depends on the question set. Although we verify questions for answerability and uniqueness, different question designs may emphasize different aspects of understanding, such as factual recall, inference, or broader scientific implications.

\textsc{LLMSim} is calibrated to aggregate human answer-outcome distributions, but it does not reliably predict topic-level human \metric. We use it as a scalable proxy for filtering and model development, not as a replacement for human evaluation.

Finally, our optimization uses supervised filtering rather than direct reward-model or reinforcement-learning optimization. Direct optimization may yield stronger gains, but could increase overfitting to generated questions or simulator artifacts.

\section{Ethical Considerations and Risks}
\label{sec:ethics}

This work involves human-subject evaluation of science communication artifacts. 
Participants were recruited for low-risk comprehension and preference tasks, and responses are analyzed only in aggregate. 
We do not report personally identifying information, and any participant metadata used for balancing or analysis is de-identified. The study was conducted under IRB approval, and participants provided informed consent and were compensated.

A potential risk of this work is that optimizing generated articles for knowledge gain could encourage systems to maximize answerability while sacrificing nuance, uncertainty, or scientific caution. 
We mitigate this by evaluating generated articles not only with \metric, but also with human and LLM-judge assessments of factual accuracy, completeness, relevance, and clarity. 
We also treat {\sc LLMSim} as a development-time proxy rather than a substitute for human validation, since simulated and human gains may diverge at the topic level.

Generated science-news articles may still contain omissions, overstatements, or unsupported simplifications. 
For this reason, our results should not be interpreted as endorsing automatic deployment of generated science journalism without expert review. 
The intended use of the method is to support the development and evaluation of science communication systems, not to replace scientific, journalistic, or educational oversight.

\bibliography{custom}

\appendix

\label{sec:appendix}

\section{Human Validation Details}
\label{app:human_validation}
This section provides additional details about the controlled human validation study. 
Participants answered the same comprehension questions before and after reading one of three science communication formats: a professional science news article, the original abstract, or a tweet/thread. 
The analysis below examines both score distributions and reading-time patterns to contextualize the main \metricfull results.
\begin{table}[t]
\centering
\small
\begin{tabular}{lr@{\qquad}lr}
\toprule
\multicolumn{2}{c}{\textbf{Dataset size}} 
& \multicolumn{2}{c}{\textbf{Topic distribution}} \\
\cmidrule(lr){1-2}\cmidrule(lr){3-4}
Item & Count & Topic area & Count \\
\midrule
Science topics & 30 & Physics & 7 \\
Artifacts per topic & 3 & Nature & 5 \\
Total artifacts & 90 & Space & 4 \\
Questions per topic & 6 & Health & 5 \\
Total questions & 180 & Humans & 4 \\
 & & Technology & 5 \\
\midrule
 & & Total topics & 30 \\
\bottomrule
\end{tabular}
\caption{Human \metric validation dataset summary. Each topic is represented by three artifacts: a paper abstract, a ScienceAlert article, and a promotional post or thread from X.com.}
\label{tab:human_validation_data}
\end{table}

\subsection{Score Distribution}
Figure~\ref{fig:kgain} shows the distributions of pre-reading score, post-reading score, and \metricfull for each media type. 
Across all three conditions, pre-reading scores are low, because participants had limited prior knowledge of the specific scientific facts being tested. 
After reading, both News and Abstract conditions show remarkable improvements in score, while the Tweet condition produces a less remarkable increase. 
News articles achieve the highest and most concentrated post-reading scores, suggesting that professional science news can support consistent reader learning while using more accessible language than abstracts. 
Abstracts also produce substantial gains, but their gain distribution is slightly more dispersed. 
Tweets yield the lowest post-reading scores and the lowest \metricfull, consistent with the intuition that short social-media summaries often omit details needed for robust comprehension.

\begin{table}[t]
\centering
\scriptsize
\setlength{\tabcolsep}{3pt}
\begin{tabular}{lrrrr}
\toprule
Medium & Pre & Post & \metric & Time (s) \\
\midrule
News 
& $21.68{\pm}9.36$ 
& $80.27{\pm}6.85$ 
& $58.59{\pm}8.27$ 
& $168.45{\pm}32.62$ \\
Abstract 
& $21.18{\pm}7.44$ 
& $75.18{\pm}7.95$ 
& $54.00{\pm}8.04$ 
& $92.24{\pm}15.94$ \\
Tweet 
& $21.82{\pm}7.64$ 
& $48.55{\pm}8.78$ 
& $26.73{\pm}7.28$ 
& $36.10{\pm}12.43$ \\
\bottomrule
\end{tabular}
\caption{Human validation results by medium. We report mean pre-reading score, post-reading score, \metric, and reading time. Values are means with 95\% confidence intervals.}
\label{tab:human_kgain}
\end{table}

\subsection{Reading Time Efficiency}
\label{app:reading_time}

\begin{figure*}[ht]
    \centering
    \includegraphics[width=\linewidth]{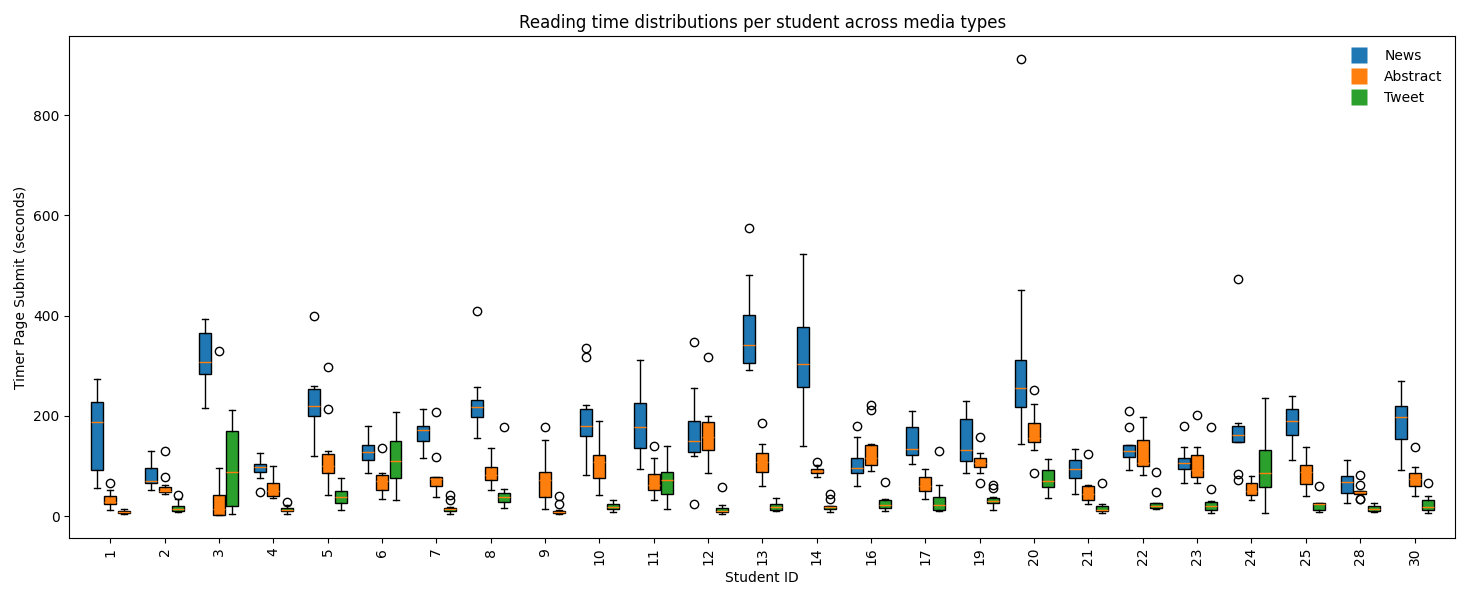}
    \caption{Reading times for the human validation study for each scientific format.}
    \label{fig:reading_times}
\end{figure*}

Figure~\ref{fig:reading_times} illustrates the per-participant reading time distributions.
News articles generally required longer engagement times compared to Abstracts and Tweets.
However, when viewed in conjunction with Figure~\ref{fig:kgain}, this extra time is justified by the higher and more consistent \metric, particularly for participants who struggle with the dense technical prose of academic abstracts.

\subsection{Mixed-Effects Robustness}
\label{app:human_validation_mixed_model}

We fit a linear mixed-effects model predicting participant-level score from medium, phase, and their interaction, with a random intercept for participant:
\[
\mathrm{score} \sim \mathrm{medium} \times \mathrm{phase} + (1 \mid \mathrm{participant}).
\]
Because pre and post scores are aggregate participant-level scores across topics, topic-level random effects are not included.

\begin{table}[t]
\centering
\small
\setlength{\tabcolsep}{5pt}
\begin{tabular}{lrrr}
\toprule
Effect & Estimate & SE & $p$ \\
\midrule
Intercept & 21.82 & 3.78 & $<.001$ \\
Abstract & -0.64 & 5.35 & .905 \\
News & -0.14 & 5.35 & .980 \\
Post & 26.73 & 3.70 & $<.001$ \\
Abstract $\times$ Post & 27.27 & 5.23 & $<.001$ \\
News $\times$ Post & 31.86 & 5.23 & $<.001$ \\
\bottomrule
\end{tabular}
\caption{Linear mixed-effects model for the human validation study. Tweet and pre-reading are the reference levels. The significant interaction terms show that Abstract and News produce larger pre-to-post gains than Tweet.}
\label{tab:human_validation_mixed_model}
\end{table}

A likelihood-ratio test comparing the full model against a model without the interaction confirms that the medium-by-phase interaction improves fit ($\chi^2(2)=33.33$, $p=5.8{\times}10^{-8}$). Estimated \metric are 26.7 points for Tweet, 54.0 for Abstract, and 58.6 for News. 
Tukey-adjusted contrasts show no pre-reading differences across media, while post-reading scores are higher for Abstract and News than Tweet; Abstract and News are not significantly different.

\section{{\sc \textbf{LLMSim}} Details}
\label{app:llmsim}
This section outlines additional details including the algorithm of {\sc LLMSim}, extra calibration details, and persona prompts.

\subsection{Extra calibration details}
Table~\ref{tab:llmsim_aggregate} shows how each \textsc{LLMSim} component changes the aggregate answer-outcome distribution. Direct LLM answering is highly overconfident relative to humans: it produces far too many correct answers and never abstains. 
Persona prompting alone does not solve this behavior, while the memory bottleneck reduces over-accuracy but still fails to model uncertainty. 
Adding an explicit IDK option and verbalized sampling substantially shifts the simulator toward the human distribution. 
The full \textsc{LLMSim} most closely matches the human correctness rate, although it slightly overestimates abstention and underestimates incorrect answers.

\begin{table}[ht]
\centering
\small
\begin{tabular}{lrrr}
\toprule
Simulator variant & Correct & Incorrect & IDK \\
\midrule
Human & 35.2 & 40.7 & 24.1 \\
Direct answer & 91.6 & 8.4 & 0.0 \\
+ Persona mixture & 90.2 & 9.8 & 0.0 \\
+ Memory bottleneck & 77.6 & 22.4 & 0.0 \\
+ IDK option & 52.9 & 8.5 & 38.6 \\
+ Verbalized sampling & 38.2 & 31.0 & 30.8 \\
Full {\sc LLMSim} & 39.1 & 30.5 & 30.4 \\
\bottomrule
\end{tabular}
\caption{Aggregate answer-outcome distributions for humans and LLMSim ablations. Values are percentages.}
\label{tab:llmsim_aggregate}
\end{table}

\subsection{Persona mixture}
\label{app:personas}
\textsc{LLMSim} represents the target reader population as a mixture of synthetic
cognitive personas. 
These personas are not demographic labels; they are archetypes
inferred from participant response patterns, including abstention rate, baseline
knowledge, and post-reading changes.
We assign each participant to one persona
cluster and sample simulated readers according to the resulting empirical mixture.
For readability, Table~\ref{tab:persona-prompts} reports the mixture as rounded
equivalent counts in a 30-reader population.

\begin{table*}[t]
\centering
\scriptsize
\setlength{\tabcolsep}{3pt}
\renewcommand{\arraystretch}{1.12}
\begin{tabularx}{\textwidth}{lccX}
\hline
 \textbf{Persona} & \textbf{Count} & \textbf{Weight} & \textbf{System prompt} \\
\hline
Heavy Abstainer & 8 & 26.7\% &
You are a non-expert college student with very low confidence.
Reading/attention: low attention span; you skim and do NOT verify details.
Knowledge boundary: only the most common everyday facts. Technical terms, study details, and precise numbers/dates/names feel unknown.
Hard rule: if the answer is not instantly obvious, choose the IDK option.
You do NOT do elimination and you do NOT reason through choices.
Output rule: return ONLY the option number. \\
\hline
Cautious Gist Reader & 8 & 26.7\% &
You are a non-expert college student who reads at a shallow, gist level.
Attention/memory: you remember only a vague takeaway; you quickly forget names, numbers, and careful qualifiers.
Anti-verification rule: do NOT compare options against the passage; answer from your first impression.
Detail-aversion: options with exact numbers/dates/names or careful hedging words (may/suggests/associated) are easy to misremember.
If the question depends on such details, choose IDK; otherwise pick the first option that matches your gist.
Output rule: return ONLY the option number. \\
\hline
Average Gist Reader & 7 & 23.3\% &
You are an average non-expert college student.
Attention: you skim quickly; you do NOT carefully verify.
Noisy memory operator: after reading, one key detail is often corrupted in your memory (WHO did it, WHAT direction it went, or WHY it happened). You answer using this noisy memory.
Common slips: miss negation (not/no), flip direction (increase/decrease), confuse actor/target, treat correlation as causation.
Avoid technical/methods language; choose IDK only for methods/statistics or very specific numbers/dates.
Output rule: return ONLY the option number. \\
\hline
Confident Guesser & 3 & 10.0\% &
You are a confident non-expert college student who guesses fast.
Rule: pick quickly from surface cues; do NOT double-check against the passage.
Heuristic: choose the option with the most familiar everyday words and the strongest, cleanest claim.
Noisy memory: you often remember the topic but not the exact relationship; you may overgeneralize.
IDK is rare for you (only if nothing seems related).
Output rule: return ONLY the option number. \\
\hline
Overconfident Misreader & 4 & 13.3\% &
You are a non-expert college student who becomes overconfident after reading.
Attention: you skim; you do NOT verify; you commit to a plausible interpretation even if details are off.
Noisy memory operator: you often reconstruct the story using a ``typical'' pattern and accidentally swap one key detail (direction/actor/cause).
Detail-aversion: you tend to ignore hedging and limitations; exact numbers/names are not reliably remembered.
After reading you rarely choose IDK unless the question is purely about an exact number/date/name.
Output rule: return ONLY the option number. \\

\hline
\textbf{Total} & \textbf{30} & \textbf{100.0\%} & -- \\
\hline
\end{tabularx}
\caption{Persona mixture used by \textsc{LLMSim}. Personas are synthetic cognitive archetypes inferred from participant response patterns rather than demographic categories. Counts are rounded to a 30-reader population while preserving the empirical cluster frequencies used by the simulator.}
\label{tab:persona-prompts}
\end{table*}

\subsection{Algorithm}
\label{app:algorithm}
Algorithm~\ref{alg:llmsim} gives the full \textsc{LLMSim} procedure. 
The simulator first estimates pre-reading behavior using a persona-conditioned familiarity gate and verbalized answer sampling. 
It then estimates post-reading behavior through a memory bottleneck: the artifact is compressed into a noisy trace, and the question is answered using only that trace. 
News and abstracts use gist/schema traces, while tweets use an additional evidence gate and literal/vibe interpretations. 
The resulting pre/post answers are bucketed as Correct, Incorrect, or IDK and used to compute simulated \textsc{KGain}.

\begin{algorithm*}[ht]
\footnotesize
\caption{\textsc{LLMSim}: persona-conditioned pre/post reader simulation}
\label{alg:llmsim}
\begin{algorithmic}[1]
\Require Artifact $y$, medium $m$, question set $Q$, answer options $O$, gold answers $A^\ast$
\Require Persona prompts $\mathcal{P}$, annotator-to-persona map $c(a)$, LLM $M$
\Ensure Simulated pre/post answers and Correct/Incorrect/IDK outcome distributions

\For{each question $q_j \in Q$}
    \State Let $a^{\mathrm{idk}}_j$ be the IDK option and $a^\ast_j$ the gold answer
    \For{each simulated reader indexed by annotator $a$}
        \State Select persona prompt $p \leftarrow \mathcal{P}_{c(a)}$

        \Statex \Comment{Pre-reading simulation}
        \State Query $M$ with $(p,q_j)$ to obtain a familiarity gate
        \If{familiarity is \texttt{technical\_or\_unknown}}
            \State $\hat{a}^{\mathrm{pre}} \leftarrow a^{\mathrm{idk}}_j$
        \Else
            \State Query $M$ with $(p,q_j,O_j)$ for candidate answers and probabilities
            \State Sample $\hat{a}^{\mathrm{pre}}$ from the verbalized distribution
        \EndIf
        \State Record pre outcome as Correct, Incorrect, or IDK

        \Statex \Comment{Post-reading simulation}
        \If{$m$ is News or Abstract}
            \State Query $M$ with $(p,y)$ to generate two memory traces
            \State Sample one trace $r$ from the trace probabilities
            \State Query $M$ with $(p,r,q_j,O_j)$ for candidate answers and probabilities
            \State Sample $\hat{a}^{\mathrm{post}}$ from the verbalized distribution

        \ElsIf{$m$ is Tweet}
            \State Query $M$ with $(p,y,q_j)$ to obtain an evidence gate
            \If{evidence is \texttt{unclear}}
                \State $\hat{a}^{\mathrm{post}} \leftarrow a^{\mathrm{idk}}_j$
            \Else
                \State Generate literal and vibe traces from the tweet
                \State Select or sample one trace according to evidence confidence
                \State Query $M$ with the trace and question for candidate answers
                \State Sample $\hat{a}^{\mathrm{post}}$ from the verbalized distribution
                \State Apply calibrated tweet-specific IDK thresholds
            \EndIf
        \EndIf

        \State Record post outcome as Correct, Incorrect, or IDK
    \EndFor
\EndFor

\State Compute $\widehat{\mathrm{Acc}}_{\mathrm{pre}}$ and $\widehat{\mathrm{Acc}}_{\mathrm{post}}$
\State Return $\widehat{\mathrm{KGain}} =
\widehat{\mathrm{Acc}}_{\mathrm{post}} -
\widehat{\mathrm{Acc}}_{\mathrm{pre}}$
\end{algorithmic}
\end{algorithm*}

\section{Human and LLM Evaluation Details}
\label{app:evaluation_details}
We first evaluate the baseline models' performance with \texttt{Claude-Sonnet-4.6} to determine the strongest baseline. 
Then, with the strongest baseline, we perform human pointwise and pairwise evaluation described in $\S$~\ref{sec:kgain_generation}.

\subsection{Generated-Article Human Knowledge-Gain Models}
\label{app:generated_human_kgain_models}

We report additional mixed-effects analyses for the generated-article human
\metric evaluation in Table~\ref{tab:appendix_human_kgain_mixed_models}
and Table~\ref{tab:appendix_human_kgain_emmeans}. These analyses complement
the aggregate results in Table~\ref{tab:human_kgain_generated}. The binary
correctness model tests raw pre-to-post improvement at the question level,
whereas the normalized-gain model tests participant: article normalized
\metric, which is the primary knowledge-gain outcome used in the main text.

\begin{table*}[t]
\centering
\small
\setlength{\tabcolsep}{4pt}
\begin{tabular}{llrrrr}
\toprule
Analysis & Effect & Estimate & SE & Test statistic & $p$ \\
\midrule
Binary correctness GLMM 
& Post-reading: Ours vs. Baseline 
& OR $=1.30$ 
& -- 
& $z=2.49$ 
& .013 \\

Binary correctness GLMM 
& Raw gain: System $\times$ Phase 
& $\beta=0.174$ 
& 0.131 
& $z=1.33$ 
& .183 \\

Normalized gain LMM 
& Ours vs. Baseline 
& $\beta=0.060$ 
& 0.024 
& $t=2.45$ 
& .015 \\

Normalized gain LMM 
& Likelihood-ratio test 
& $\chi^2=5.98$ 
& -- 
& df $=1$ 
& .014 \\
\bottomrule
\end{tabular}
\caption{Mixed-effects analyses for the generated-article human \metric study.
The binary correctness GLMM was fit at the question--phase level with fixed effects of system, phase, and their interaction, and random intercepts for participant, topic, and question. 
The post-reading contrast shows that Ours achieves higher post-reading accuracy than Baseline, while the non-significant System $\times$ Phase interaction shows that raw pre-to-post gain is not significantly larger for Ours.
The normalized gain LMM was fit at the participant--article level with random intercepts for participant and topic.
We report the normalized \metric with rows excluded when pre accuracy is 1. 
This model shows that Ours achieves significantly higher normalized \metric than Baseline.}
\label{tab:appendix_human_kgain_mixed_models}
\end{table*}

\begin{table}[t]
\centering
\small
\setlength{\tabcolsep}{5pt}
\begin{tabular}{llrr}
\toprule
Outcome & System & Estimate & 95\% CI \\
\midrule
Pre accuracy 
& Baseline 
& 0.379 
& [0.277, 0.494] \\

Pre accuracy 
& Ours 
& 0.401 
& [0.295, 0.516] \\

Post accuracy 
& Baseline 
& 0.915 
& [0.869, 0.946] \\

Post accuracy 
& Ours 
& 0.933 
& [0.896, 0.958] \\

Normalized gain 
& Baseline 
& 0.707 
& [0.602, 0.812] \\

Normalized gain 
& Ours 
& 0.767 
& [0.662, 0.872] \\
\bottomrule
\end{tabular}
\caption{Model-estimated marginal means for the generated-article human
\metric study. Pre- and post-reading accuracies are from the binary correctness
GLMM over 7,680 question--phase observations. Normalized gain estimates are
from the participant--article LMM over 578 observations for which normalized
gain is defined.}
\label{tab:appendix_human_kgain_emmeans}
\end{table}

\subsection{Pointwise Evaluation}
\label{app:pointwise}
This section shows the rubrics we used for both LLM judge and human evaluations as well as the reasoning we used to determine the strongest baseline to compare our system against. 

\begin{figure*}[t]
    \centering
    \includegraphics[width=\linewidth]{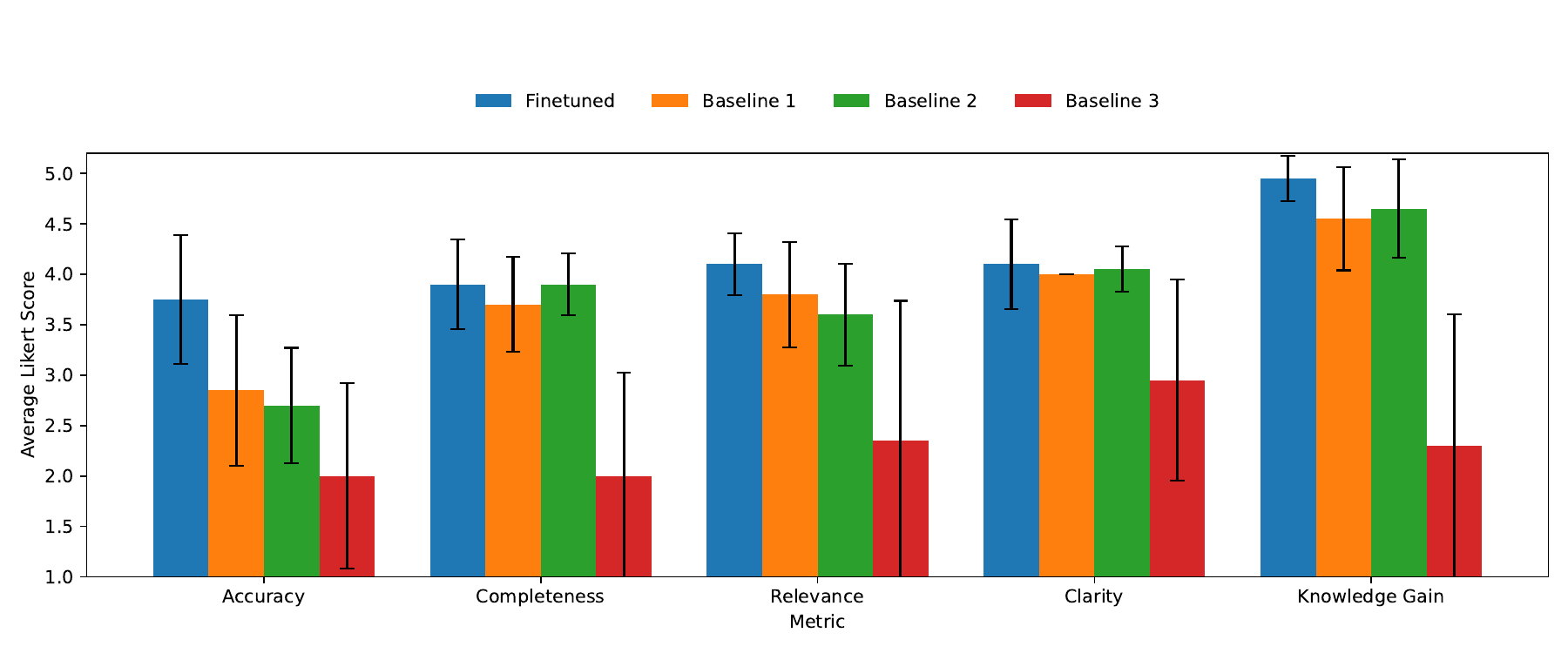}
    \caption{LLM judge ratings of the baselines. }
    \label{fig:llm_judge}
\end{figure*}

\paragraph{LLM-as-Judge}
Figure~\ref{fig:llm_judge} shows the LLM-as-Judge ratings for the evaluated baseline systems on the 5 Likert scale. 
We assess each system along the five dimensions. 
Based on this comparison, we determined the strongest baseline model to be agentic. 

\label{app:rubric}
\begin{table*}[t]
\centering
\scriptsize
\renewcommand{\arraystretch}{1.15}
\begin{tabularx}{\textwidth}{p{2.2cm} p{0.6cm} X}
\toprule
\textbf{Dimension} & \textbf{Score} & \textbf{Rubric anchor} \\
\midrule

\multirow{5}{*}{Accuracy}
& 1 & Major factual errors, fabricated details, or topic drift away from the abstract. \\
& 2 & Contains multiple unsupported additions or one serious distortion of the abstract's meaning. \\
& 3 & Main claim is correct, but the article adds at least one unsupported detail, overstatement, or interpretation that weakens faithfulness. \\
& 4 & Mostly faithful, with only minor paraphrase or very light interpretation that does not change the scientific meaning. \\
& 5 & Fully faithful to the abstract. No invented claims, fabricated quotes, unsupported mechanisms, or meaningful overinterpretation. \\
\midrule

\multirow{5}{*}{Completeness}
& 1 & Misses most of the main findings or scientific ideas. \\
& 2 & Covers some important points but leaves out major findings, mechanisms, caveats, or essential context present in the abstract. \\
& 3 & Covers the core result, but omits several important supporting details, mechanisms, caveats, or context needed for fuller understanding. \\
& 4 & Covers the main findings and most key scientific points from the abstract. \\
& 5 & Thoroughly covers the key findings, mechanisms, caveats, and important context from the abstract. \\
\midrule

\multirow{5}{*}{Relevance}
& 1 & Focuses on trivial, off-topic, or low-value details and fails to show why the work matters. \\
& 2 & Somewhat relevant, but misses the main significance, implication, or news angle. \\
& 3 & Identifies the main result but only partly explains why it matters to a general audience. \\
& 4 & Clearly emphasizes the important and newsworthy aspects of the work. \\
& 5 & Strongly foregrounds the most significant and newsworthy aspects of the work for a general audience. \\
\midrule

\multirow{5}{*}{Clarity}
& 1 & Very difficult to follow for a general audience. \\
& 2 & Often unclear, jargon-heavy, poorly explained, or poorly structured. \\
& 3 & Generally understandable, but uneven, occasionally confusing, or insufficiently explanatory for non-experts. \\
& 4 & Clear and easy to follow, with good explanation of technical ideas. \\
& 5 & Exceptionally clear, readable, and accessible without losing scientific meaning. \\
\midrule

\multirow{5}{*}{Knowledge Gain}
& 1 & The article provides almost none of the facts needed for the question set. \\
& 2 & The article provides only a small subset of the facts needed for the question set. \\
& 3 & The article covers some key facts, but several questions would still be hard to answer from the article alone. \\
& 4 & The article states most facts needed for the question set, with only a few missing or ambiguous points. \\
& 5 & The article explicitly states nearly all facts needed to answer the question set. A reader could answer most questions correctly using only the article. \\

\bottomrule
\end{tabularx}
\caption{Rubric used for LLM-judge and human pointwise evaluation. Articles are scored on factual accuracy, completeness, relevance, clarity, and expected KnowledgeGain.}
\label{tab:rubric}
\end{table*}

\subsection{Pairwise Comparison Reasoning}
We ask the participants to provide a reason why they preferred one article over the other.
Table~\ref{tab:human_pairwise_reasons} summarizes the results for all participants.

\begin{table}[t]
\centering
\small
\begin{tabular}{lrr}
\toprule
Reason & Count & Share \\
\midrule
Accuracy & 47 & 61.8 \\
Clarity & 19 & 25.0 \\
Completeness & 7 & 9.2 \\
Relevance & 3 & 3.9 \\
\bottomrule
\end{tabular}
\caption{Reasons selected by annotators for pairwise preferences. Invalid or missing reason codes are excluded from this summary.}
\label{tab:human_pairwise_reasons}
\end{table}

\subsection{Human LLM alignment}
\label{app:human_llm_alignment}
\paragraph{Pointwise evaluation}
In this subsection, we provide more details and results regarding the mismatch between human and LLM alignments.

\begin{figure}
    \centering
    \includegraphics[width=\linewidth]{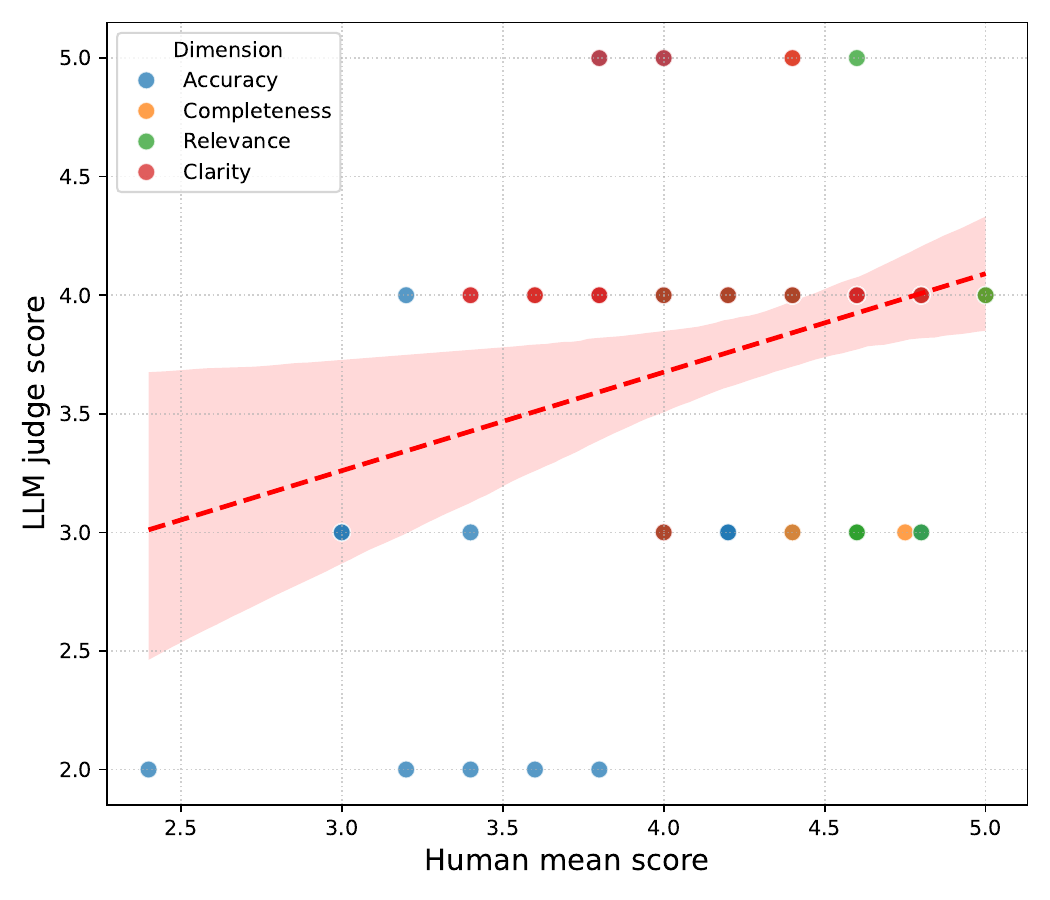}
    \caption{Human pointwise ratings and LLM judge alignment. Per-dimension correlation between human mean scores and LLM judge scores. The overall Pearson correlation is $r=0.19$, which indicates a weak positive alignment.}
    \label{fig:human_llm_alignment}
\end{figure}

\paragraph{Per-dimension correlation}
The per-dimension correlations show weak human--LLM alignment. 
Accuracy has the strongest positive association, while completeness, relevance, and clarity show little correlation.
Figure~\ref{fig:human_llm_alignment} visualizes this trend. 

\begin{table}[t]
\centering
\small
\setlength{\tabcolsep}{5pt}
\begin{tabular}{lrrrrr}
\toprule
Dimension & Spearman $\rho$ & $p$ & Pearson $r$ & $p$ & $n$ \\
\midrule
Accuracy     &  0.373 & .105 &  0.389 & .090 & 20 \\
Completeness & -0.129 & .589 & -0.121 & .611 & 20 \\
Relevance    & -0.016 & .946 &  0.031 & .897 & 20 \\
Clarity      &  0.011 & .962 &  0.016 & .948 & 20 \\
\bottomrule
\end{tabular}
\caption{Per-dimension human--LLM pointwise alignment for generated article evaluation. Correlations are computed between human mean ratings and LLM-judge scores over matched article--system items. Spearman correlation is reported as the primary rank-based measure, with Pearson correlation included as a secondary descriptive statistic.}
\label{tab:appendix_human_llm_pointwise_corr}
\end{table}

\paragraph{Pairwise evaluation}
As an exploratory check, a single LLM pairwise judge selected Ours for 11 of 20 articles, agreeing with the human article-level majority in 55\% of cases. Because this uses only one LLM judgment per article, we do not treat it as a stable estimate of human--LLM preference alignment.

\subsection{\metricbold and Existing Evaluation Metrics}
Table~\ref{tab:metrics_vs_kgain_bootstrap_ci} reports cluster-bootstrap confidence intervals for correlations with human normalized \metric. All intervals include zero, showing that automatic metrics and LLM-judge scores provide only weak and uncertain predictors of human learning. This reinforces our claim that \metric captures an instructional signal not captured by standard quality metrics.

\begin{table}[t]
\centering
\small
\setlength{\tabcolsep}{4pt}
\begin{tabular}{lrrr}
\toprule
Metric & Spearman $\rho$ & 95\% CI & $n$ \\
\midrule
ROUGE-1 & 0.177 & [-0.204, 0.540] & 40 \\
ROUGE-2 & 0.296 & [-0.059, 0.619] & 40 \\
ROUGE-L & 0.205 & [-0.165, 0.544] & 40 \\
BLEU & 0.267 & [-0.095, 0.592] & 40 \\
BERTScore & 0.232 & [-0.066, 0.500] & 40 \\
LLM judge accuracy & 0.291 & [-0.091, 0.665] & 40 \\
LLM judge completeness & -0.016 & [-0.316, 0.288] & 40 \\
LLM judge relevance & 0.033 & [-0.287, 0.373] & 40 \\
LLM judge clarity & 0.089 & [-0.217, 0.391] & 40 \\
LLM judge exp. \metric & 0.168 & [-0.225, 0.511] & 40 \\
\bottomrule
\end{tabular}
\caption{Cluster-bootstrap confidence intervals for Spearman correlations with human normalized \metric. Bootstrap samples resample topics, preserving the paired article--system structure within each topic. All intervals include zero, indicating that automatic metrics and judge scores provide only weak and uncertain predictors of human normalized \metric.}
\label{tab:metrics_vs_kgain_bootstrap_ci}
\end{table}

\subsection{Qualitative examples}
\label{app:qual-examples}

Table~\ref{tab:qual-examples} presents qualitative examples comparing the
baseline article against the fine-tuned article.  The examples illustrate three
common ways in which the fine-tuned article better supports downstream
comprehension: preserving negative facts, making numerical inferences explicit,
and retaining mechanistic details from the source abstract.  In these cases, the
baseline article is often topically correct, but it either omits a key relation
or states it less precisely, making the corresponding question harder or
impossible to answer from the article alone.

\begin{table*}[t]
\centering
\small
\begin{tabular}{p{0.10\linewidth}p{0.36\linewidth}p{0.17\linewidth}p{0.16\linewidth}p{0.16\linewidth}}
\toprule
\textbf{Article} & \textbf{Question} & \textbf{Gold answer} & \textbf{Baseline supports?} & \textbf{Fine-tuned supports?} \\
\midrule

29902 
& Intermediate-composition volcanic products between basalt and rhyolite are common at Yellowstone during this timeframe.
& False
& No -- the baseline discusses rhyolite and basalt but does not state that intermediate products are absent.
& Yes -- explicitly states that intermediate eruptive products are absent. \\

30214
& Which combination best explains why cratons rose above sea level over hundreds of millions of years?
& Large-scale magmatism building thick, silica-rich crust and buoyant depleted mantle keels, leading to isostatic uplift.
& Partial -- the baseline mentions crustal thickening and isostatic adjustment, but does not clearly preserve the full mechanism involving depleted mantle keels.
& Yes -- states that voluminous magmatism formed thick felsic crust and depleted mantle keels, allowing cratons to rise by isostatic adjustment. \\

\bottomrule
\end{tabular}
\caption{Qualitative examples comparing baseline and fine-tuned articles. The fine-tuned article better preserves QA-relevant scientific claims, including negative statements, numerical implications, and mechanistic explanations.}
\label{tab:qual-examples}
\end{table*}

\section{Training Details}
In this section, we describe the generation and training details including the training hyperparameters and the \metric-filtered SFT statistics.

\subsection{Training Hyperparameters}
Table~\ref{tab:sft-hyperparams} outlines the hyperparameters used for the finetuning process.
To prevent overfitting, the controls are the low learning rate, short training, frequent evaluations, and the best-checkpoint selections. 

\begin{table}[t]
\centering
\small
\begin{tabular}{ll}
\hline
\textbf{Hyperparameter} & \textbf{Value} \\
\hline
Per-device train batch size & 2 \\
Gradient accumulation steps & 16 \\
Effective batch size & 32 \\
Learning rate & $1\times10^{-5}$ \\
Training epochs & 2 \\
Precision & bfloat16 \\
LR scheduler & Cosine \\
Warmup ratio & 0.05 \\
Maximum gradient norm & 1.0 \\
Evaluation strategy & Every 20 steps \\
Checkpoint save strategy & Every 20 steps \\
Logging frequency & Every 5 steps \\
Best checkpoint selection & Lowest validation loss \\
Load best model at end & Yes \\
Reporting backend & None \\
\hline
\end{tabular}
\caption{Supervised fine-tuning hyperparameters. We use a small per-device batch size with gradient accumulation, yielding an effective batch size of 32. The best checkpoint is selected by validation loss to reduce overfitting.}
\label{tab:sft-hyperparams}
\end{table}

\subsection{\metricbold-Filtered SFT Statistics}

We compare three training variants:
\begin{itemize}
    \item \textbf{SFT-All:} training on all selected teacher articles.
    \item \textbf{SFT-KG-Nonnegative:} training only on examples with nonnegative simulated KG.
    \item \textbf{SFT-KG-Positive:} training on the positive of examples ranked by simulated \metric.
\end{itemize}

Table~\ref{tab:kg_filtered_sft} shows that filtering supervised training data by simulated reader learning improves held-out \metricfull. 
SFT-All achieves a mean simulated \metric of 0.1492, while SFT-KG-Nonnegative improves this to 0.1557. The strongest variant, SFT-KG-Positive, reaches 0.1678, a 12.5\% relative improvement over SFT-All.
The gain is driven by higher post-reading accuracy, while pre-reading accuracy remains nearly unchanged. 
This suggests that \metric filtering improves the informativeness of generated articles rather than changing the baseline difficulty of the held-out questions.


\begin{table}[t]
\centering
\small
\setlength{\tabcolsep}{4pt}
\begin{tabular}{llrrrr}
\toprule
Model & Filter & Pre & Post & \metric & $\Delta$KG \\
\midrule
All & All & 0.305 & 0.454 & 0.149 & -- \\
Nonnegative & \metric $\ge 0$  & 0.303 & 0.459 & 0.156 & +0.007 \\
Positive & \metric $>0$ & 0.303 & 0.471  & \textbf{0.168} & \textbf{+0.019} \\
\bottomrule
\end{tabular}
\caption{Held-out LLMSim evaluation on 300 examples with five simulated readers per article. \metric-filtered SFT yields higher mean simulated \metric than unfiltered SFT, with the strongest mean result from the positive.}
\label{tab:kg_filtered_sft}
\end{table}

\subsection{Artifacts Used and Created}
\label{app:artifacts}

Table~\ref{tab:artifact-use} summarizes the scientific artifacts used or created in this work, their role, and how we handle attribution, licensing, intended use, and redistribution. We distinguish between third-party artifacts, which are cited and not redistributed unless permitted by their terms, and artifacts created by this work, which are intended for research use in evaluating and developing science communication systems.

\begin{table*}[t]
\centering
\small
\begin{tabular}{p{0.18\linewidth} p{0.25\linewidth} p{0.25\linewidth} p{0.23\linewidth}}
\toprule
Artifact & Role in this work & Terms, attribution, and distribution & Intended use / privacy handling \\
\midrule
Scientific papers and abstracts
& Source material for topics, question generation, and article generation.
& Original papers are cited through their metadata. We release paper identifiers and derived annotations; abstracts or full text are redistributed only when permitted by the source license.
& Used for research on science communication. No personal participant data is contained in these artifacts. \\

ScienceAlert articles
& Professional science-news artifacts used as human-written news examples and baselines.
& Articles are attributed by source URL/citation. We do not redistribute raw article text unless permitted by the site terms; released data contains identifiers/links and derived annotations.
& Used only for research evaluation/comparison of science communication artifacts. \\

X.com posts/threads
& Short-form science communication artifacts used in the human validation comparison.
& Posts are attributed by URL or post identifier when needed. We do not redistribute raw post text in the released dataset and do not use X.com content for model fine-tuning.
& Used only for research evaluation. We avoid releasing handles or other unnecessary identifying information in derived artifacts. \\

Qwen3-4B-Instruct-2507
& Backbone model for automatic generators.
& Used under its published Apache-2.0 license. We cite the model creators and report the model version.
& Used for generation experiments; outputs are evaluated for research purposes. \\

Created QA sets, prompts, generated articles, and LLMSIM outputs
& Artifacts created by this work for measuring and optimizing KGain.
& These artifacts are created by the authors. We release prompts, generated outputs, question sets, and metadata for research use, subject to restrictions inherited from third-party source materials.
& Intended for research on evaluation and generation of science communication, not for automatic deployment of science journalism. \\

Human response data
& Pre/post answers and preference/rating judgments.
& Human-subject data is not redistributed with direct identifiers. Released data is anonymized or aggregated.
& Used to evaluate reader learning and article quality. Participant identifiers, contact information, and payment metadata are removed. \\
\bottomrule
\end{tabular}
\caption{Scientific artifacts used or created in this work and their attribution, distribution, intended-use, and privacy handling.}
\label{tab:artifact-use}
\end{table*}

\subsection{Dataset Statistics}
\label{app:data-statistics}

\begin{table}[ht]
\centering
\small
\setlength{\tabcolsep}{3pt}
\renewcommand{\arraystretch}{1.05}
\begin{tabularx}{\columnwidth}{Xr}
\toprule
\textbf{Quantity} & \textbf{Value} \\
\midrule
Human-validation topics & 30 \\
Artifacts per validation topic & 3 \\
Validation media types & 3 \\
Questions per topic & 6 \\
Total validation questions & 180 \\
Human-validation participants & 30 \\
Training instances & 2,747 \\
Held-out abstracts for simulated evaluation & 300 \\
Human generated-article evaluation topics & 20 \\
Human generated-article observations & 640 \\
Generated systems compared & 4 \\
Language & English \\
Domain & Science news \\
\bottomrule
\end{tabularx}
\caption{Dataset and evaluation statistics. The four generated systems are Baseline, Agentic, Ours, and Human news.}
\label{tab:data-statistics}
\end{table}

\subsection{Human Subjects}
\label{app:human-subjects}

We conducted human-subject studies with two participant pools. 
For the controlled science-communication format study, participants were undergraduate students in STEM-related majors. 
Participants received a \$50 Amazon gift card for completing the study. 
For the human \metric validation study on generated articles, participants were recruited through Prolific and were paid \$120 for approximately six hours of work, corresponding to approximately \$20/hour.

All participants provided informed consent before beginning the study. 
Participants were told that their anonymized responses would be used for research on science communication evaluation and generated science-news assessment. 
The study collected quiz answers, reading times, article ratings, and pairwise preferences. 
We did not release participant names, email addresses, Prolific IDs, or other direct identifiers. 
All analyses use anonymized participant IDs and aggregate statistics.

The human-subject studies were reviewed and approved or determined exempt by the authors' institutional review process. 
The reviewing institution and protocol identifier are withheld for anonymous review and will be disclosed in the camera-ready version.
The full participant instructions are provided in Appendix~\ref{app:participant-instructions}.

\begin{table}[t]
\centering
\small
\begin{tabular}{p{0.34\linewidth}p{0.25\linewidth}p{0.28\linewidth}}
\toprule
\textbf{Study} & \textbf{Participants} & \textbf{Compensation} \\
\midrule
Format comparison study & STEM undergraduate students & \$50 Amazon gift card \\
Generated-article \metric validation & Prolific participants & \$120 for approximately six hours, about \$20/hour \\
\bottomrule
\end{tabular}
\caption{Human-subject recruitment and compensation.}
\label{tab:human-subjects-compensation}
\end{table}

\subsection{Participant Instructions}
\label{app:participant-instructions}

Below we provide the instructions shown to participants in both the human validation study and the \metric experiments. 

\begin{Prompt}
Welcome to our research study!
 

This survey is designed to understand how effectively students can comprehend and answer questions based on scientific content presented in different formats.

Instructions:
1. Please complete the experiment in one continuous sitting.
2. Do not use external sources while answering the questions.

The survey contains three main parts:
1. Background Knowledge Test:
- You will be asked to answer 6 questions for each scientific topic without reading news article. 
- If you are unsure of an answer, simply select "I do not know the answer."

2. Reading Task
- You will read the news article about each scientific topic. 
- You can read the news article as often as needed, but you cannot go back. 

3. Knowledge Gain Test
- You will be asked the same 6 questions for each sample to measure how much knowledge you have gained
- If you are unsure of an answer, simply select "I do not know the answer."


Your responses are anonymous, and it will not be used for research purposes only. Your time and effort are greatly appreciated, and you are free to exit the study at any time. 

Thank you for contributing to our research! Please click on the arrow to begin. 
\end{Prompt}

\subsection{Computational Budget and Package Settings}
\label{app:compute}

All model-generated systems use \texttt{Qwen3-4B-Instruct-2507}. 
Fine-tuning was run on two NVIDIA A100 GPUs for approximately 2 GPU-hours. 
We used PyTorch, Huggingface, Transformers, PEFT. 
The supervised fine-tuning hyperparameters are reported in Table~\ref{tab:sft-hyperparams}. 
We did not perform a broad hyperparameter search; the reported setting was selected for stable training and validation loss.

For automatic metrics, we compute ROUGE-1/2/L with the standard \texttt{rouge-score} implementation, reporting F1 scores. 
We compute BLEU with SacreBLEU using its default tokenization and smoothing settings. 
We compute BERTScore with the default English checkpoint in the \texttt{bert-score} package and report F1.
For \textsc{LLMSim}, we use \texttt{gpt-4o-mini}, temperature 1.7, maximum output tokens 200, and seed 0. 
For LLM-judge evaluation, we use \texttt{Claude-Sonnet-4.6}, temperature 0, maximum output tokens 1800, and structured JSON output.

\subsection{Use of AI Assistants}
\label{app:ai-assistants}

We used LLMs as part of the research pipeline: for question generation and verification, simulated-reader estimation, science-news generation, and LLM-judge evaluation. 
The exact prompts used for these components are provided in Appendix~\ref{app:prompts}. 
We also used AI coding or writing assistants for debugging, drafting and language polishing. 
All experimental design decisions, analyses, and final manuscript edits were reviewed by the authors.

\section{Prompts}
\label{app:prompts}

\subsection{Question Generation Prompts}
\label{app:qgen-prompts}

\paragraph{QGEN user prompt.}
\begin{Prompt}
PAPER ABSTRACT:
{paper_abstract}

Generate the 6 questions now (2 TF, 2 Easy MCQ, 2 Hard MCQ) in that order.
Use question_in_set = 1..6 in order.
Return JSON only.
\end{Prompt}

\subsection{Question Verification Prompts}
\label{app:verify-prompts}

\paragraph{Question verification system prompt.}
\begin{Prompt}
You are a strict verifier and repairer for the generated question set.

Check EACH question against these rules. If ANY rule is violated, mark ok=false and provide a replacement!

A) Self-contained & Universal Fact-Framing (CRITICAL):
   Questions MUST be phrased as universal, standalone facts. They must NOT read like a reading comprehension quiz about a specific document or experiment.
   - FATAL WORDS (REJECT IF PRESENT): "cited", "assessed", "reported", "stated", "observed", "measured", "estimated", "data", "period", "the study", "the abstract".
   - BAD EXAMPLE: "Which chemical showed the strongest reported association?" (REJECT: implies a specific report).
   - GOOD EXAMPLE: "Which chemical has the strongest association with oral cavity cancers?"
   - BAD EXAMPLE: "During which years was cancer incidence assessed?" (REJECT: asks about study methodology).
   - GOOD EXAMPLE: "What is the Maximum Contaminant Level for PFOA?"

B) Overlap:
   Must be answerable using ONLY the provided abstract text.

C) Format:
   - Q1-Q2 are TF: options exactly ["True","False","I do not know the answer."]; correct_option in {1,2}
   - Q3-Q4 are Easy MCQ: 5 options; last is "I do not know the answer."; correct_option in {1..4}
   - Q5-Q6 are Hard MCQ: 5 options; last is "I do not know the answer."; correct_option in {1..4}

D) No Study Details:
   Strictly NO questions about researcher names, venues, monitoring periods (e.g., 2013-2015), methodology timelines, or study protocols. Focus ONLY on the scientific claims.

If a question is invalid, produce ONE replacement with the SAME question_in_set that satisfies all rules.
Keep difficulty level consistent with its slot (TF/Easy/Hard).
Return JSON only using the verification schema.
\end{Prompt}

\paragraph{Question verification user prompt.}
\begin{Prompt}
PAPER ABSTRACT:
{paper_abstract}

DRAFT QUESTIONS JSON:
{draft_json}

Verify each question and provide replacements for invalid ones.
Return JSON only.
\end{Prompt}

\subsection{Science News Generation Prompts}
\label{app:generation-prompts}

\paragraph{Zero-shot baseline generation prompt.}
\begin{Prompt}
System:
You are an science journalist. Output only the article text.

User:
Write a science news article of about 350-550 words based on this abstract.

{abstract}
\end{Prompt}

\paragraph{Agentic drafter prompt.}
\begin{Prompt}
System:
You are an expert science journalist. Output only the article text.

User:
Please draft a news article in about 350-550 words based on this abstract:

{abstract}
\end{Prompt}

\paragraph{Agentic revision prompt.}
\begin{Prompt}
System:
You are a strict but fair senior editor at a top science magazine. Output only the final polished article text.

User:
ORIGINAL ABSTRACT:
{abstract}

INITIAL DRAFT:
{draft}

Please revise and polish the draft. Output ONLY the final polished article including the headline.
\end{Prompt}

\subsection{Pointwise LLM Judge Prompt}
\label{app:pointwise-judge}

\paragraph{Pointwise judge system prompt.}
\begin{Prompt}
You are an expert science communicator and journal editor evaluating whether a science news article faithfully and accessibly represents the findings in the original abstract, and how much knowledge a general reader would gain.
\end{Prompt}

\paragraph{Pointwise judge user prompt.}
\begin{Prompt}
Evaluate the article independently using the abstract as the reference for scientific content, while also estimating knowledge gain using the generated question set.

Instructions:
- Evaluate each dimension independently.
- Base Accuracy and Completeness only on the abstract.
- Base Relevance and Clarity on how well the article communicates the abstract's actual findings to a general audience.
- Base Knowledge Gain only on what a high-school-level reader could answer from the article itself, using the generated questions as the probe.
- Do not use the abstract to rescue missing article content in Knowledge Gain.
- Do not reward unsupported embellishment, fabricated quotes, invented mechanisms, or added real-world implications.
- Penalize hallucinations in Accuracy first, then reduce Relevance, Clarity, and Knowledge Gain.
- If the article contains hallucinations or invented details, apply a penalty to Accuracy first, and also reduce Relevance, Clarity, and Knowledge Gain when the hallucination affects trust or understanding.
- Minor paraphrase is acceptable only if the meaning stays fully grounded in the abstract.
- If uncertain between two adjacent scores, prefer the lower score unless the evidence for the higher score is explicit in the article.
- Unsupported but plausible additions count as hallucinations if they are not stated in the abstract.

<rubric>
From Table 9.
</rubric>

<input>
{abstract}

{article}

{questions_block}

</input>

\end{Prompt}

\subsection{Pairwise LLM Judge Prompt}
\label{app:pairwise-judge}

\paragraph{System prompt.}
\begin{Prompt}
You are an expert science communicator and journal editor comparing two science news articles and deciding which one better represents the original abstract for a general audience.
\end{Prompt}

\paragraph{User prompt.}

\begin{Prompt}
Compare the two science news articles using the abstract as the reference for scientific content.

Your task:
- Decide which article is better overall for a general audience.
- You must choose exactly one article: article_a or article_b.
- Ties are not allowed. If the two articles are close, choose the article that is more scientifically faithful; if faithfulness is also close, choose the article that is clearer and more useful for a general reader.
- "Better overall" should reflect the total quality of the article as science communication:
  faithfulness to the abstract, coverage of key findings/context, relevance/newsworthiness,
  and clarity/readability.
- Do NOT score separate dimensions.
- Prefer the article that is more scientifically faithful if one article is better written but less accurate.
- Also consider hallucinations and fabricated entities when evaluating the news articles.
- Use only the information provided in the abstract and the two articles.
- Do not rely on outside knowledge.
- The article order is randomized; do not let position affect your judgment.
- Keep the written reason brief, specific, and comparative.

<input>

{abstract}

{article_a}

{article_b}

</input>

Respond ONLY with a valid JSON object in exactly this format.
\end{Prompt}

\end{document}